\documentclass{IEEEtran}
\usepackage[T1]{fontenc}
\usepackage{color}
\usepackage{array}
\usepackage{booktabs}
\usepackage{url}
\usepackage{multirow}
\usepackage{amsmath}
\usepackage{amssymb}
\usepackage{graphicx}
\usepackage{wasysym}
\usepackage{xargs}[2008/03/08]
\PassOptionsToPackage{normalem}{ulem}
\usepackage{ulem}
\usepackage[unicode=true,
 bookmarks=true,bookmarksnumbered=false,bookmarksopen=false,
 breaklinks=false,pdfborder={0 0 1},backref=false,colorlinks=false]
 {hyperref}

\makeatletter

\providecommand{\tabularnewline}{\\}

\usepackage{graphicx}%Required
\usepackage{url}
\usepackage{colortbl} 
\definecolor{header_color}{rgb}{0.70,0.70,0.70}
\definecolor{even_color}{rgb}{0.9,0.9,0.9}
\definecolor{subheader_color}{rgb}{0.85,0.85,0.85}
\definecolor{childheader_color}{rgb}{1.0,0.93,0.87}
\makeatother

\begin{document}
\title{OptiGAN: Generative Adversarial Networks for Goal Optimized Sequence
Generation}
\author{\author{\IEEEauthorblockN{Mahmoud Hossam\IEEEauthorrefmark{1}, Trung Le\IEEEauthorrefmark{1}, Viet Huynh\IEEEauthorrefmark{1}, Michael Papasimeon\IEEEauthorrefmark{2} and Dinh Phung\IEEEauthorrefmark{1}}\\ \IEEEauthorblockA{\IEEEauthorrefmark{1}Faculty of Information Technology, Monash University, Australia}\\ \IEEEauthorblockA{\IEEEauthorrefmark{2}School of Computing and Information Systems, The University of Melbourne, Australia} \\  \small{\texttt{\IEEEauthorrefmark{1}\{mhossam, trunglm, viet.huynh, dinh.phung\}@monash.edu, \IEEEauthorrefmark{2}michael.papasimeon@unimelb.edu.au}}}}

\maketitle
\newcommand{\sidenote}[1]{\marginpar{\small \emph{\color{Medium}#1}}}

\global\long\def\se{\hat{\text{se}}}%

\global\long\def\interior{\text{int}}%

\global\long\def\boundary{\text{bd}}%

\global\long\def\ML{\textsf{ML}}%

\global\long\def\GML{\mathsf{GML}}%

\global\long\def\HMM{\mathsf{HMM}}%

\global\long\def\sup{\text{sup}}%

\global\long\def\new{\text{*}}%

\global\long\def\stir{\text{Stirl}}%

\global\long\def\mA{\mathcal{A}}%

\global\long\def\mB{\mathcal{B}}%

\global\long\def\mF{\mathcal{F}}%

\global\long\def\mK{\mathcal{K}}%

\global\long\def\mL{\mathcal{L}}%

\global\long\def\mH{\mathcal{H}}%

\global\long\def\mX{\mathcal{X}}%

\global\long\def\mZ{\mathcal{Z}}%

\global\long\def\mS{\mathcal{S}}%

\global\long\def\Ical{\mathcal{I}}%

\global\long\def\mT{\mathcal{T}}%

\global\long\def\Pcal{\mathcal{P}}%

\global\long\def\dist{d}%

\global\long\def\HX{\entro\left(X\right)}%
 
\global\long\def\entropyX{\HX}%

\global\long\def\HY{\entro\left(Y\right)}%
 
\global\long\def\entropyY{\HY}%

\global\long\def\HXY{\entro\left(X,Y\right)}%
 
\global\long\def\entropyXY{\HXY}%

\global\long\def\mutualXY{\mutual\left(X;Y\right)}%
 
\global\long\def\mutinfoXY{\mutualXY}%

\global\long\def\given{\mid}%

\global\long\def\gv{\given}%

\global\long\def\goto{\rightarrow}%

\global\long\def\asgoto{\stackrel{a.s.}{\longrightarrow}}%

\global\long\def\pgoto{\stackrel{p}{\longrightarrow}}%

\global\long\def\dgoto{\stackrel{d}{\longrightarrow}}%

\global\long\def\lik{\mathcal{L}}%

\global\long\def\logll{\mathit{l}}%

\global\long\def\vectorize#1{\mathbf{#1}}%

\global\long\def\vt#1{\mathbf{#1}}%

\global\long\def\gvt#1{\boldsymbol{#1}}%

\global\long\def\idp{\ \bot\negthickspace\negthickspace\bot\ }%
 
\global\long\def\cdp{\idp}%

\global\long\def\das{:=}%

\global\long\def\id{\mathbb{I}}%

\global\long\def\idarg#1#2{\id\left\{  #1,#2\right\}  }%

\global\long\def\iid{\stackrel{\text{iid}}{\sim}}%

\global\long\def\bzero{\vt 0}%

\global\long\def\bone{\mathbf{1}}%

\global\long\def\boldm{\boldsymbol{m}}%

\global\long\def\bff{\vt f}%

\global\long\def\bx{\boldsymbol{x}}%

\global\long\def\bl{\boldsymbol{l}}%

\global\long\def\bu{\boldsymbol{u}}%

\global\long\def\bo{\boldsymbol{o}}%

\global\long\def\bh{\boldsymbol{h}}%

\global\long\def\bs{\boldsymbol{s}}%

\global\long\def\bz{\boldsymbol{z}}%

\global\long\def\xnew{y}%

\global\long\def\bxnew{\boldsymbol{y}}%

\global\long\def\bX{\boldsymbol{X}}%

\global\long\def\tbx{\tilde{\bx}}%

\global\long\def\by{\boldsymbol{y}}%

\global\long\def\bY{\boldsymbol{Y}}%

\global\long\def\bZ{\boldsymbol{Z}}%

\global\long\def\bU{\boldsymbol{U}}%

\global\long\def\bv{\boldsymbol{v}}%

\global\long\def\bn{\boldsymbol{n}}%

\global\long\def\bV{\boldsymbol{V}}%

\global\long\def\bI{\boldsymbol{I}}%

\global\long\def\bw{\vt w}%

\global\long\def\balpha{\gvt{\alpha}}%

\global\long\def\bbeta{\gvt{\beta}}%

\global\long\def\bmu{\gvt{\mu}}%

\global\long\def\btheta{\boldsymbol{\theta}}%

\global\long\def\blambda{\boldsymbol{\lambda}}%

\global\long\def\bgamma{\boldsymbol{\gamma}}%

\global\long\def\bpsi{\boldsymbol{\psi}}%

\global\long\def\bphi{\boldsymbol{\phi}}%

\global\long\def\bpi{\boldsymbol{\pi}}%

\global\long\def\bomega{\boldsymbol{\omega}}%

\global\long\def\bepsilon{\boldsymbol{\epsilon}}%

\global\long\def\btau{\boldsymbol{\tau}}%

\global\long\def\realset{\mathbb{R}}%

\global\long\def\realn{\realset^{n}}%

\global\long\def\integerset{\mathbb{Z}}%

\global\long\def\natset{\integerset}%

\global\long\def\integer{\integerset}%

\global\long\def\natn{\natset^{n}}%

\global\long\def\rational{\mathbb{Q}}%

\global\long\def\rationaln{\rational^{n}}%

\global\long\def\complexset{\mathbb{C}}%

\global\long\def\comp{\complexset}%

\global\long\def\compl#1{#1^{\text{c}}}%

\global\long\def\and{\cap}%

\global\long\def\compn{\comp^{n}}%

\global\long\def\comb#1#2{\left({#1\atop #2}\right) }%

\global\long\def\nchoosek#1#2{\left({#1\atop #2}\right)}%

\global\long\def\param{\vt w}%

\global\long\def\Param{\Theta}%

\global\long\def\meanparam{\gvt{\mu}}%

\global\long\def\Meanparam{\mathcal{M}}%

\global\long\def\meanmap{\mathbf{m}}%

\global\long\def\logpart{A}%

\global\long\def\simplex{\Delta}%

\global\long\def\simplexn{\simplex^{n}}%

\global\long\def\dirproc{\text{DP}}%

\global\long\def\ggproc{\text{GG}}%

\global\long\def\DP{\text{DP}}%

\global\long\def\ndp{\text{nDP}}%

\global\long\def\hdp{\text{HDP}}%

\global\long\def\gempdf{\text{GEM}}%

\global\long\def\rfs{\text{RFS}}%

\global\long\def\bernrfs{\text{BernoulliRFS}}%

\global\long\def\poissrfs{\text{PoissonRFS}}%

\global\long\def\iidrfs{$\text{IidRFS}$}%

\global\long\def\grad{\gradient}%
 
\global\long\def\gradient{\nabla}%

\global\long\def\partdev#1#2{\partialdev{#1}{#2}}%
 
\global\long\def\partialdev#1#2{\frac{\partial#1}{\partial#2}}%

\global\long\def\partddev#1#2{\partialdevdev{#1}{#2}}%
 
\global\long\def\partialdevdev#1#2{\frac{\partial^{2}#1}{\partial#2\partial#2^{\top}}}%

\global\long\def\hessian{\text{Hess}}%

\global\long\def\closure{\text{cl}}%

\global\long\def\cpr#1#2{\Pr\left(#1\ |\ #2\right)}%

\global\long\def\var{\text{Var}}%

\global\long\def\Var#1{\text{Var}\left[#1\right]}%

\global\long\def\cov{\text{Cov}}%

\global\long\def\Cov#1{\cov\left[ #1 \right]}%

\global\long\def\COV#1#2{\underset{#2}{\cov}\left[ #1 \right]}%

\global\long\def\corr{\text{Corr}}%

\global\long\def\sst{\text{T}}%

\global\long\def\SST{\sst}%

\global\long\def\ess{\mathbb{E}}%

\global\long\def\Ess#1{\mathbb{E}\left[#1\right]}%

\newcommandx\ESS[2][usedefault, addprefix=\global, 1=]{\underset{#2}{\mathbb{E}}\left[#1\right]}%

\global\long\def\fisher{\mathcal{F}}%

\global\long\def\bfield{\mathcal{B}}%
 
\global\long\def\borel{\mathcal{B}}%

\global\long\def\bernpdf{\text{Bernoulli}}%

\global\long\def\betapdf{\text{Beta}}%

\global\long\def\dirpdf{\text{Dir}}%

\global\long\def\gammapdf{\text{Gamma}}%

\global\long\def\gaussden#1#2{\text{Normal}\left(#1, #2 \right) }%

\global\long\def\gauss{\mathbf{N}}%

\global\long\def\gausspdf#1#2#3{\text{Normal}\left( #1 \lcabra{#2, #3}\right) }%

\global\long\def\multpdf{\text{Mult}}%

\global\long\def\poiss{\text{Pois}}%

\global\long\def\poissonpdf{\text{Poisson}}%

\global\long\def\pgpdf{\text{PG}}%

\global\long\def\wshpdf{\text{Wish}}%

\global\long\def\iwshpdf{\text{InvWish}}%

\global\long\def\nwpdf{\text{NW}}%

\global\long\def\niwpdf{\text{NIW}}%

\global\long\def\studentpdf{\text{Student}}%

\global\long\def\unipdf{\text{Uni}}%

\global\long\def\transp#1{\transpose{#1}}%
 
\global\long\def\transpose#1{#1^{\mathsf{T}}}%

\global\long\def\mgt{\succ}%

\global\long\def\mge{\succeq}%

\global\long\def\idenmat{\mathbf{I}}%

\global\long\def\trace{\mathrm{tr}}%

\global\long\def\argmax#1{\underset{_{#1}}{\text{argmax}} }%

\global\long\def\argmin#1{\underset{_{#1}}{\text{argmin}\ } }%

\global\long\def\diag{\text{diag}}%

\global\long\def\norm{}%

\global\long\def\spn{\text{span}}%

\global\long\def\vtspace{\mathcal{V}}%

\global\long\def\field{\mathcal{F}}%
 
\global\long\def\ffield{\mathcal{F}}%

\global\long\def\inner#1#2{\left\langle #1,#2\right\rangle }%
 
\global\long\def\iprod#1#2{\inner{#1}{#2}}%

\global\long\def\dprod#1#2{#1 \cdot#2}%

\global\long\def\norm#1{\left\Vert #1\right\Vert }%

\global\long\def\entro{\mathbb{H}}%

\global\long\def\entropy{\mathbb{H}}%

\global\long\def\Entro#1{\entro\left[#1\right]}%

\global\long\def\Entropy#1{\Entro{#1}}%

\global\long\def\mutinfo{\mathbb{I}}%

\global\long\def\relH{\mathit{D}}%

\global\long\def\reldiv#1#2{\relH\left(#1||#2\right)}%

\global\long\def\KL{KL}%

\global\long\def\KLdiv#1#2{\KL\left(#1\parallel#2\right)}%
 
\global\long\def\KLdivergence#1#2{\KL\left(#1\ \parallel\ #2\right)}%

\global\long\def\crossH{\mathcal{C}}%
 
\global\long\def\crossentropy{\mathcal{C}}%

\global\long\def\crossHxy#1#2{\crossentropy\left(#1\parallel#2\right)}%

\global\long\def\breg{\text{BD}}%

\global\long\def\lcabra#1{\left|#1\right.}%

\global\long\def\lbra#1{\lcabra{#1}}%

\global\long\def\rcabra#1{\left.#1\right|}%

\global\long\def\rbra#1{\rcabra{#1}}%

\begin{abstract}
One of the challenging problems in sequence generation tasks is the
optimized generation of sequences with specific desired goals. Current
sequential generative models mainly generate sequences to closely
mimic the training data, without direct optimization of desired goals
or properties specific to the task. We introduce OptiGAN, a generative
model that incorporates both Generative Adversarial Networks (GAN)
and Reinforcement Learning (RL) to optimize desired goal scores using
policy gradients. We apply our model to text and real-valued sequence
generation, where our model is able to achieve higher desired scores
out-performing GAN and RL baselines, while not sacrificing output
sample diversity.
\end{abstract}


\thispagestyle{empty}

\section{Introduction\label{sec:Introduction}}

Learning to generate realistic sequences from existing data is essential
to many artificial intelligence applications, including text generation,
drug design, robotics, and music synthesis. In these applications,
a generative model learns to generate sequences of different data
types according to each task. For instance, natural language and speech
are sequences of words or utterances, in robot motion planning, a
trajectory is an action sequence learned from experiences or sensory
data. Recently, there has been a growing interest in deep models
for sequence generation following the success of Generative Adversarial
Networks (GANs) \cite{goodfellow2014generative} and Variational Autoencoders
(VAEs) \cite{DBLP:journals/corr/KingmaW13} in image generation tasks
\cite{radford2015unsupervised,10.5555/3295222.3295327,zhang2018selfattention,dam2019three,razavi2019generating}.

However, realizing the full potential of these models in aforementioned
applications has many challenges, and one of these key challenges
is the absence of mechanisms to optimize the generated outputs according
to certain metrics or useful properties. Most of current work on generative
sequence models mainly learn to ``resemble'' the data, meaning to
generate outputs that are close to the real distribution. However,
in many applications, we are not only interested in generating data
similar to the real ones, but we need them to have specific useful
properties or attributes. For example, in drug design, useful properties
include solubility and ease of synthesis \cite{de2018molgan,guimaraes2017objectivereinforced}.
In music generation, we might want the music to have specific pitch
or tempo, or in text applications, the user might be interested in
generating sentences according certain sentiment or tense \cite{pmlr-v70-hu17e}.
Therefore, the lack of optimization mechanisms in current models
hinders their practical use in wide range of real world applications.\textcolor{blue}{}

In this paper, we propose a new sequential generative framework, named
OptiGAN\footnote{Our code is available here : \url{https://github.com/mahossam/OptiGAN}},
that can generate sequences resembling those in a given dataset and
achieving high scores according to an optimized goal (e.g., solubility
and ease of synthesis in drug design). Our proposed framework leverage
GAN for mimicking real data and policy gradient reinforcement learning
(RL) \cite{Williams92simplestatistical} for optimizing a score of
interest. It is very well-known that although GANs can resemble real
data, they face the mode collapsing problem \cite{goodfellow2016nips,nguyen2017dual,hoang2018mgan,le2018geometric,le2019learning},
hence leading to generate less diverse examples. To tackle this issue
in the context of sequence generation, we propose a principled combination
of maximum likelihood and GANs (see Section \ref{subsec:GAN_ML})
in which we prove that in the final optimization problem, Kullback-Leibler
(KL) and Jensen-Shannon (JS) divergences between real data distribution
and generated data distribution are simultaneously maximized, hence
relieving the mode collapsing problem, concurring with \cite{nguyen2017dual}.
We then leverage policy gradient RL into our model for optimizing
a score of interest according to a desired goal (see Section \ref{subsec:RL}).
We observe that when incorporating policy gradient RL to our current
framework -which is based on GAN and maximum likelihood- the variance
in estimating gradient is very high, hence leading to unstable training.
To resolve this issue, we resort the Monte Carlo rollout in \cite{yu2017seqgan}
with a slight modification (see Section \ref{subsec:Reducing-policy-gradients}).

We demonstrate the capacity of our OptiGAN in two applications: text
generation (discrete data) and air combat trajectory generation (real-valued
data). For text generation task, we aim to generate sentences resembling
real sentences in a given text corpus, while optimizing the BLEU \cite{DBLP:conf/acl/PapineniRWZ02}
score for obtaining better sentences from human justification. For
\textit{\emph{aircraft trajectory generation}}\textit{ }task, we aim
to generate a trajectory plan for air-combat maneuver scenario between
two aircrafts and optimize the McGrew score \cite{mcgrew2010air}
which reflects the tactic quality of aircraft trajectories in an air
combat \cite{mcgrew2010air}. In both applications, we show that we
can generate high quality outputs and achieve higher scores than current
related models aided by the RL component, while preserving the diversity
of generated outputs using our hybrid maximum likelihood GAN.\textcolor{blue}{{}
}

The main contributions of our paper include:
\begin{itemize}
\item We propose OptiGAN which has the following advantages: (i) \emph{an
end-to-end generative framework with incorporated goal optimization
mechanism,} (ii) \emph{general formulation that can be used for wide
variety of different goals and models},\textcolor{blue}{{} }and (iii)
\emph{optimizing for desired goals without sacrificing output sample
diversity}. 
\item To the best of our knowledge, OptiGAN is the first GAN controlled
generative model for sequences that addresses the diversity issue
in a principled approach. OptiGAN combines the benefits of GAN and
RL policy learning, while avoiding their mode-collapse and high variance
drawbacks. 
\item We investigate the problems of interest comprehensively and our findings
would advance the understanding of the behavior when incorporating
GAN-based models with RL. Specifically, we empirically show that if
only pure RL is applied to maximize a score of interest with the GAN-based
objective, realistic quality of the output might be sacrificed for
the sake of superficially obtaining higher scores. For instance, in
the case of text generation, the model was able to cheat the selected
quality score by generating sentences in which few words are repeated
all the time. This shows that combining a GAN-based objective with
RL encourages the optimization process of RL to stay close to the
true data distribution. 
\end{itemize}

\section{\label{sec:BG-Related-Work}Background and related work}

\subsection{\label{subsec:Background}Background}

\subsubsection*{Generative Adversarial Networks (GAN)}

Generative Adversarial Networks (GAN) \cite{goodfellow2014generative}
use adversarial training between two players to learn the density
function of input data. The goal of the first player, the generator
$G$, is to get good at generating data that is close to the real
data distribution $p_{d}(x)$. The goal of the second player, the
discriminator $D$, is to distinguish real data from fake data generated
by the generator. The standard GAN objective to optimize is the minimax
game between $D$ and $G$ is :
\begin{equation}
\min_{G}\max_{D}\left(\mathbb{{E}}_{x\sim p_{d}}\log D(x)+\mathbb{{E}}_{z\sim p_{z}}\log(1-D(G(z)))\right),
\end{equation}
where $z$ is the random noise input to $G$ and $p_{z}$ is the prior
distribution of the $z$. After the training is finished, the generator
is used to generate data from any random input $z$.

\subsubsection*{\label{subsec:Reinforcement-Learning-using}Reinforcement Learning
using Policy Gradients}

Reinforcement learning \cite{sutton2011reinforcement} is a general
learning framework in which an agent learns how to take actions to
maximize cumulative future rewards in the environment. Rewards received
by the agent during learning encourage it to learn a policy that maximizes
cumulative rewards (the returns).

Policy gradients \cite{sutton2011reinforcement} are group of methods
in reinforcement learning that enable optimizing future returns by
direct optimization of the policy. The objective is to maximize the
return rewards over an episode of $T$ time steps $\begin{aligned}J(\theta) & =\mathbb{E}_{\pi}\left[U_{t}\right]\end{aligned}
$, where $\pi$ is the ``policy'' and $U_{t}$ specifies the cumulative
reward of an episode which is defined as follows:
\[
U_{t}\doteq R_{t+1}+\gamma R_{t+2}+\gamma^{2}R_{t+3}+\cdots+\gamma^{T-t-1}R_{T},
\]
where $\gamma$ is a discount factor, and $R_{t}$ is the reward received
from the environment at time $t$.

A policy could be parameterized by some parameters $\theta$ and be
directly optimized through taking the gradient of $J(\theta)$ with
respect to $\theta$ . This method is called policy gradients. A well-known
policy gradients algorithm is REINFORCE \cite{Williams92simplestatistical},
a Monte Carlo algorithm to find the optimal policy $\pi$. The model
is updated via gradient ascent with:
\begin{equation}
\begin{aligned}\nabla J(\boldsymbol{\theta}) & =\mathbb{E}_{\pi}[\sum_{t}U_{t}\nabla_{\theta}\log\pi\left(A_{t}\mid S_{t},\theta\right)]\end{aligned}
,\label{eq:PG}
\end{equation}
where $A_{t}$ is an action chosen at time step $t$ by the agent's
policy $\pi$ given the current state of the environment $S_{t}$.

\subsection{\label{Related-Work}Related work}

In general, sequential deep generative models are either based on
the variational approximation of maximum likelihood (like Variational
Autoencoders VAE \cite{DBLP:journals/corr/KingmaW13}) or on GANs
\cite{goodfellow2014generative}. Models based on variational approximation
\cite{NIPS2015_5653,NIPS2018_7424,fraccaro2016sequential,roberts2018hierarchical}
are mainly based on autoregressive models like Long Short-Term Memory
(LSTM) \cite{hochreiter1997long}, incorporated into VAE training
framework. These models were applied to many sequence generating tasks
including handwriting and music generation. However, training VAE
based models with autoregressive networks suffers from the problem
of ``\textit{posterior collapse}'', where the latent variables
are often ignored, especially when trained for discrete data like
text \cite{bowman2016generating}.

The other group of models based on GANs are mainly focused on discrete
data like text. There are two main approaches for these models; they
either use reinforcement learning \cite{yu2017seqgan,fedus2018maskgan},
or a fully differentiable GAN \cite{nie2018relgan,chen2018adversarial,zhang2017adversarial,kusner2016gans}.
The first approach uses policy gradients in an adversarial training
framework.  The other approach however, employs a fully differentiable
GAN network, where they use Gumbel-Softmax trick \cite{maddison2016concrete,jang2016categorical}
or distance measure on feature space \cite{chen2018adversarial} to
overcome the non-differentiability problem for discrete data.

Some recent work tried to address the goal optimized generation
problem. In \cite{engel2018gansynth}, authors used convolutional
GAN or autoregressive VAE to generate music with specific pitch and
timbre. For text generation, \cite{pmlr-v70-hu17e} uses semi-supervised
VAE approach to generate text based on sentiment and tense. Interest
is growing as well in the biological sequences and drug design applications
\cite{doi:10.1021/acs.molpharmaceut.7b01137}, where VAE latent space
or GAN based model with reinforcement learning are used for molecular
design.  However, these models either optimize using a RL objective
or employ feature learning in the generative VAE or GAN model. There
is not much work on using RL to guide GAN learning for goal optimization,
combining benefits of GAN unsupervised learning with goal optimization.
MolGAN \cite{de2018molgan}, is a recent work in that direction, that
uses both GAN and RL for optimized graph generation. However, we additionally
address the diversity issue of GANs in a principled way through our
maximum likelihood GAN.

\section{Problems of interest}

We demonstrate the capacity of our proposed framework in two applications
of interest: \emph{text generation} and \emph{air combat trajectory
generation}. For each application, our task is to generate sequences
that achieve two concurrent goals: i) mimicking those in a given dataset
and ii) obtaining high scores specified by an optimized goal which
might be varied for specific tasks.

\subsection{Text generation \label{subsec:Text-generation}}

We need to generate sentences that are similar to real sentences in
a given text corpus and have high quality from human justification.
A well-known score used to measure the quality of generated sentences
is BLEU score \cite{DBLP:conf/acl/PapineniRWZ02}. Specifically, the
BLEU score for each sentence computes the ratio of n-grams generated
from the model that matches with a true ground truth, called \textit{reference
}sentences and is defined as follows:\begin{equation}\resizebox{1\linewidth}{!}{$\textrm{BLEU-N}=\prod_{n=1}^{N}\left(\textrm{precision}_{n}=\frac{Count(\text{ Model}\text{ }n\text{-grams}\,\cap X_{\textrm{ref.}n\text{-grams}})}{Count(\text{ Model}\text{ }n\text{-grams})}\right)^{\textrm{weight}(n)}$}\end{equation}

In our proposed model, beside generating realistic sentences, we
also aim to maximize the BLEU score of generated sentences. As shown
later, we utilize the BLEU score as reward function in our RL inspired
framework. 

\subsection{Air combat trajectory generation}

\begin{figure}
\begin{centering}
\includegraphics[width=0.99\columnwidth]{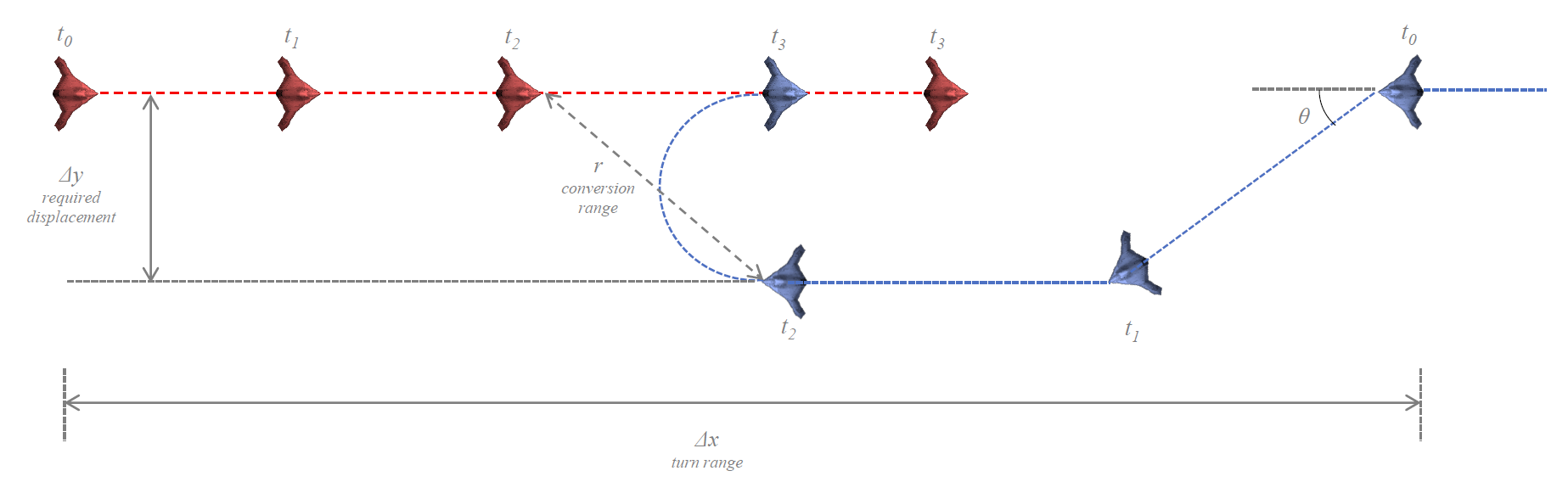}
\par\end{centering}
\caption{\label{fig:Stern-Conversion-Maneuvre}\textquotedblleft Stern Conversion\textquotedblright{}
Flight Maneuver.}
\end{figure}

For air combat missions, pilots are trained to conduct certain maneuvers
according the combat situation they face. There are well known maneuvers
that the pilots are trained on, either defensive, offensive, or neutral.
We consider a specific air combat maneuver between two fighters called
``\textbf{Stern Conversion}'' maneuver \cite{austin1990game}. In
this maneuver, the opponent (the red aircraft) flies in a straight
and level line, and does not detect the blue aircraft, while the blue
aircraft, on the other hand, tries to get behind the opponent aircraft,
in order to increase the chance to engage it (see Fig. \ref{fig:Stern-Conversion-Maneuvre}).

In this specific task, in addition to generating realistic trajectories,
we also need to maximize the McGrew score \cite{mcgrew2010air}, which
measures the score of how well an aircraft was doing relative to another
aircraft in an attempt to get behind the other aircraft (refer to
\cite{mcgrew2010air} for more detail). Due to security restrictions,
we cannot access the real trajectories sensory data. Instead, we use
\textit{ACE-Zero \cite{ijcai2017-778} }air combat flight simulator
to generate the training data. This \textit{simulator }\textit{\emph{was
developed by domain experts to imitate the real aircraft trajectories.}}\textit{
}\textit{\emph{ As demonstrated in the experiments section, our model
can generate novel trajectories with high McGrew score close to the
average scores for ACE-Zero trajectories.}}

\section{\label{sec:Proposed-Framework}Proposed framework}

\begin{figure*}
\begin{centering}
\includegraphics[viewport=60bp 30bp 900bp 530bp,clip,width=0.6\textwidth]{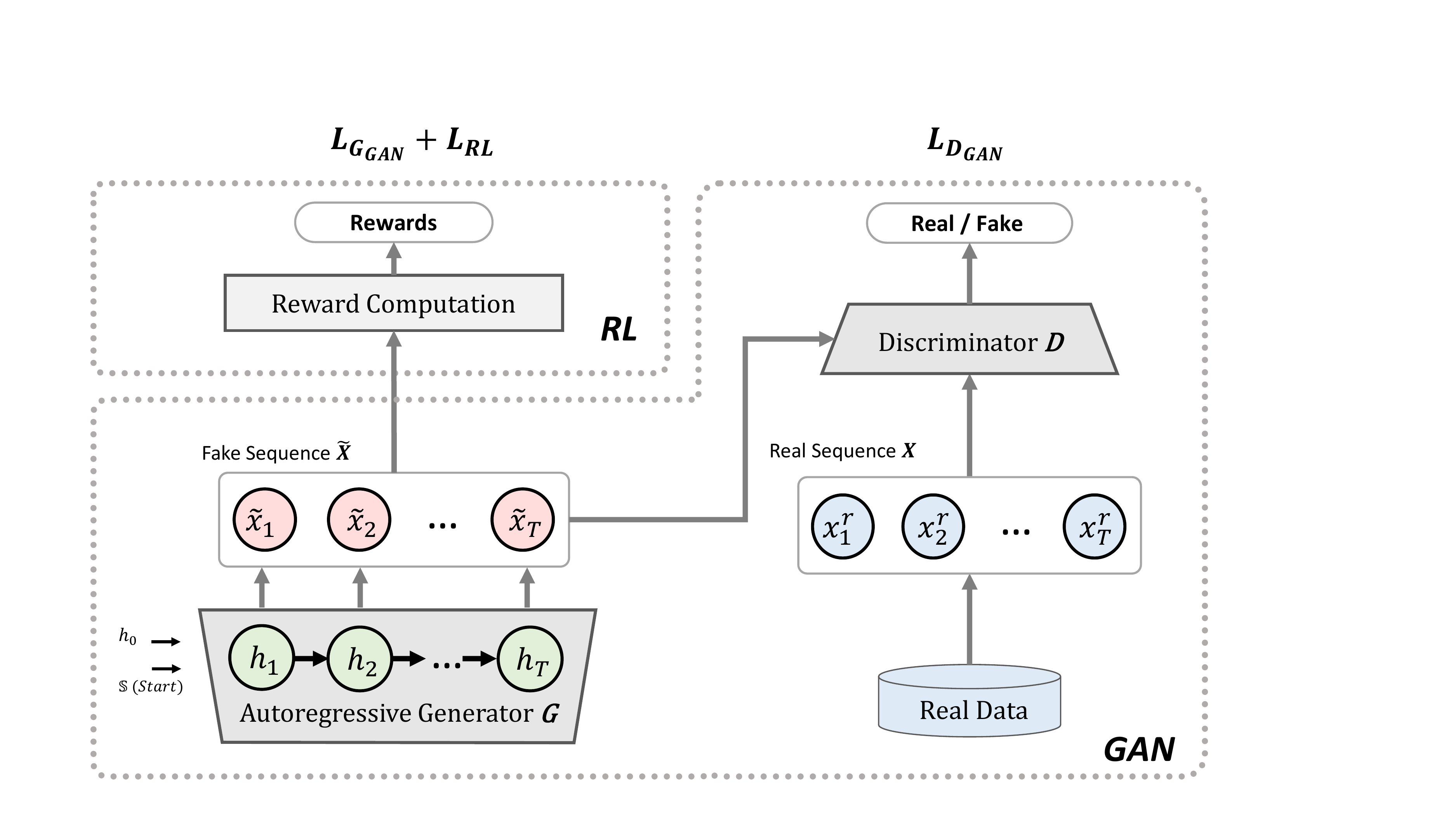}
\par\end{centering}
\centering{}\caption{Overview of OptiGAN framework. The Reinforcement Learning (RL) component
is incorporated with sequence GAN model. The generator $G$ is trained
by combining two losses, the GAN loss and the RL loss, $L_{G_{GAN}}$
and $L_{RL}$.\label{fig:Model}}
\end{figure*}

\textbf{}In what follows, we present in details our proposed framework.
We employ a neural autoregressive model $G$ (e.g., Bi-RNN or RNN)
as a generator to map from a noise $z\sim p_{z}$ to a sequence that
can mimic those in a given dataset and achieve high score corresponding
to the desired goal. In terms of modeling, we start from the maximum
likelihood (ML) principle and then propose to incorporate adversarial
learning to the learning process in a principled way. The coupling
of ML and adversarial learning principles helps us to generate realistic
and diverse sequences to imitate those in a given dataset. Moreover,
to reach high scores according to a given desired goal, we propose
to incorporate policy gradient RL that allows us to train our model
end-to-end. Finally, to stabilize the training process, we apply variance
reduction technique when training with policy gradients. The final
model is named OptiGAN whose overview architecture is shown in Fig.
\ref{fig:Model}.

\subsection{Maximum likelihood and adversarial training \label{subsec:GAN_ML}}

A sample $X$ in our setting is defined as a sequence of $T$ tokens
denoted by $X=\left[x_{1},x_{2},...,x_{T}\right]$, where we assume
that all samples have length $T$. For our autoregressive model with
model parameters $\theta$, the log-likelihood can be written as:
\begin{align*}
\log\,p_{G}(X\mid\theta) & =\sum_{i=2}^{T}\log\,p_{G}(x_{i}\mid h_{i-1},\theta)+\log\,p_{G}(x_{1}\mid\theta),
\end{align*}
This is the default neural autoregressive model formulation. Now we
start introducing an adversarial learning framework for this model
by introducing a latent variable $z$ to the autoregressive model,
where we rewrite $\log\,p(x_{1}\mid\theta)$ as marginalization over
the $z$:{\small{}
\begin{gather}
\log\,p_{G}(x_{1}\mid\theta)=\log\,\sum_{z}\,p_{G}(x_{1},z\mid\theta)\geqslant\label{eq:1}\\
\,\,\,\,\,\,\,\,\,\,\,-I_{KL}(q(z\mid x_{1},\phi)\,||\,p(z))+\mathbb{E}_{q(z\mid x_{1},\phi)}[\log\,p_{G}(x_{1}\mid z,\theta)],\nonumber 
\end{gather}
}where $I_{KL}$ is Kullback--Leibler divergence, $q(z\mid x_{1},\phi)$
is an approximation of the posterior $p\left(z\mid x_{1},\theta\right)$
and $p(z)$ is a prior distribution to $z$. The right hand side of
Eq. (\ref{eq:1}) is a lower bound for $\log\,p_{G}(x_{1}\mid\theta)$.
We can then write $\log\,p_{G}(X\mid\theta)$ in terms of a lower
bound as:{\small{}
\begin{align}
\log\,p_{G}(X & \mid\theta)\geqslant\sum_{i=2}^{T}\log\,p_{G}(x_{i}\mid h_{i-1},\theta)\label{eq:2}\\
 & -I_{KL}(q(z\mid x_{1},\phi)\,\Vert\,p(z))+\mathbb{E}_{q(z\mid x_{1},\phi)}[\log\,p_{G}(x_{1}\mid z,\theta)].\nonumber 
\end{align}
}{\small\par}

We propose to incorporate adversarial learning to autoregressive
sequential model in a principled way. One generator $G\left(z\right)$
and one discriminator $D\left(X\right)$ are employed to create a
game like in GAN while the task of the discriminator is to discriminate
true data and fake data and the task of the generator is to generate
fake data that maximally make the discriminator confused. In addition,
the generator $G$ is already available which departs from a noise
$z\sim p_{z}$, uses the conditional distribution $p\left(x_{1}\mid z,\theta\right)$
to generate $x_{1}$, and follows the autoregressive model to consecutively
generate $x_{2:T}$. We come with the following minimax problem:\textbf{}\\
\begin{align}
\max_{G}\min_{D} & (\mathbb{E}_{X\sim p_{d}}\left[\log\,p_{G}\left(X\mid\theta\right)\right]-\text{\ensuremath{\mathbb{E}}}_{X\sim p_{d}}\left[\log\,D\left(X\right)\right]\nonumber \\
 & \,\,\,\,\,\,\,\,\,\,\,\,\,\,\,\,\,\,\,-\mathbb{E}_{z\sim p_{z}}\left[\log\left[1-D\left(G\left(z\right)\right)\right]\right]),\label{eq:op_ARN}
\end{align}
where the generator $G$ consists of the decoder $p\left(x_{1}\mid z,\theta\right)$,
the autoregressive model, hence $G$ is parameterized by $\left(\theta,\phi\right)$,
and $\log\,p_{G}\left(X\mid\theta\right)$ is substituted by its lower
bound in Eq. (\ref{eq:2}).We can theoretically prove that the minimax
problem in Eq. (\ref{eq:op_ARN}) is equivalent to the following optimization
problem (see the proof in Appendix \ref{subsec:Proof-of-final-1}):
\begin{equation}
\min_{G}\,I_{KL}\left(P_{d}\,||\,P_{G}\right)+I_{JS}\left(P_{d}\,||\,P_{G}\right),\label{eq:op_diver}
\end{equation}
where $I_{JS}$ is Jenshen-Shannon divergence and $P_{G}$ is the
generative distribution. The optimization problem in Eq. (\ref{eq:op_diver})
reveals that at the Nash equilibrium point the generative distribution
$P_{G}$ is exactly the data distribution $P_{d}$, thus overcoming
the mode-collapse issue caused by original GAN formulation \cite{nguyen2017dual}.

To train our model, we alternatively update $G$ and $D$ with relevant
terms. We note that in the optimization for updating $G$ regarding
$\log\,p_{G}\left(X\mid\theta\right)$, we maximize its lower bound
in Eq. (\ref{eq:2}) instead of the likelihood function.

\textbf{Training procedure. }To train our model, we alternatively
update the discriminator and generator:
\begin{itemize}
\item Update $D$:{\small{}
\[
\max_{D}\,\text{\ensuremath{\mathbb{E}}}_{X\sim p_{d}}\left[\log\,D\left(X\right)\right]+\mathbb{E}_{z\sim p_{z}}\left[\log\left[1-D\left(G\left(z\right)\right)\right]\right].
\]
}{\small\par}
\item Update $G$:{\small{}
\begin{gather}
\max_{G}\,\mathbb{E}_{X\sim p_{d}}\left[\log\,p_{G}\left(X\mid\theta\right)\right]-\mathbb{E}_{z\sim p_{z}}\left[\log\left[1-D\left(G\left(z\right)\right)\right]\right]\nonumber \\
=\max_{G}\mathbb{E}_{X\sim p_{d}}\left[\log\,p_{G}\left(X\mid\theta\right)\right]+\mathbb{E}_{z\sim p_{z}}\left[\log\,D\left(G\left(z\right)\right)\right].\label{eq:update_G}
\end{gather}
}{\small\par}
\end{itemize}
It is worth noting that for discrete data (e.g. text), we define the
likelihood $p_{G}(x_{i}\mid h_{i})=\text{softmax}(W_{o}h_{i})$ where
$W_{o}$ is the output weight matrix. In addition, to allow end to
end training, we apply Gumbel softmax \cite{maddison2016concrete,jang2016categorical}
trick for the discrete case, and fix start token to $p(x_{1}|z)=0$,
as we depend on Gumbel Softmax for random output sampling. For real-valued
data (e.g., air combat trajectory), we employ $p_{G}(x_{i}\mid h_{i})=\mathcal{N}\left(W_{o}h_{i},\sigma^{2}\right)$
where $\sigma$ is the standard deviation parameter.

\subsection{Optimizing score corresponding to a goal with reinforcement learning
\label{subsec:RL}}

To incorporate the ability to model the data to maximize rewards from
the environment, we use policy gradient to learn a policy that maximizes
the total rewards from environment. Following \cite{sutton2011reinforcement},
the learning objective to maximize the return rewards over an episode
from $t=[0,\,1,...,\,T-1]$ is:
\[
\begin{aligned}J(\boldsymbol{\theta}) & =\mathbb{E}_{\pi}[\sum_{t}U_{t}\log\pi\left(A_{t}|S_{t},\theta\right)]\end{aligned}
,
\]

\noindent where $\pi$ is the ``policy'', or the probability distribution
of actions given states of environment, $A_{t}$ and $S_{t}$ are
the action and state at time $t$, $\theta$ are the parameters of
$\pi$ and $U_{t}$ is ``the return rewards'' at time $t$. 

In our model, the policy is the generator $G$, and the state at time
$t$ is the hidden state of the generator $h_{t}$. Thus the objective
becomes:\\
\[
\begin{aligned}J(\theta) & =\mathbb{E}_{X\sim P_{G}}[\sum_{t}U_{t}\log p_{G}\left(X\mid h_{t},\theta\right)]\end{aligned}
.
\]

We then use REINFORCE \cite{Williams92simplestatistical} algorithm
to find the optimum parameters for policy $G$ by gradient ascent
of the gradient of $J$ as:

\[
\nabla J(\theta)=\mathbb{E}_{X\sim P_{G}}[\sum_{t}U_{t}\nabla_{\theta}\log p_{G}\left(X\mid h_{t},\theta\right)].
\]

\noindent  The choice of $U_{t}$ depends on the task. For example,
we use the BLEU score as the reward value for the text generation
task, while for the air combat trajectory generation task, we use
the McGrew score as the reward.

\textbf{Total loss. }The final total loss to train the generator $G$
with adversarial training and policy gradients is:

\begin{align*}
\max_{G}\mathbb{E}_{X\sim p_{d}}\log p_{G}\left(X\mid\theta\right)+\mathbb{\lambda\,E}_{z\sim p_{z}}\log D\left(G\left(z\right)\right)\\
+\alpha\mathbb{E}_{X\sim P_{G}}[\sum_{t}U_{t}\log p_{G}\left(X\mid h_{t},\theta\right)]
\end{align*}

\noindent where $\lambda$ and $\alpha$ are hyper-parameters that
control how much the effect of adversarial training and policy gradients
are on the total loss.

\subsection{Reducing policy gradients variance \label{subsec:Reducing-policy-gradients}}

In order to reduce the variance of policy gradients, we use an algorithm
similar to the Monte Carlo rollout in \cite{yu2017seqgan} with a
slight modification. We generate few complete sentences at each time
step onward, and take their average as $U_{t}$ at that time step.
Instead of getting the reward value from a discriminator as in \cite{yu2017seqgan},
we directly compute the reward according to the chosen score.

In addition, to further reduce the variance of the policy gradients
and to help policy gradients converge faster toward optimal solution,
we use policy gradients with baseline \cite{sutton2011reinforcement},
where policy gradient is defined as:
\begin{align*}
\nabla J(\boldsymbol{\theta}) & =\mathbb{E}_{\pi}[\sum_{t}\left(U_{t}-b\left(S_{t}\right)\right)\nabla_{\theta}\log\pi\left(A_{t}\mid S_{t},\theta\right)]\\
= & \mathbb{E}_{X\sim P_{G}}[\sum_{t}\left(U_{t}-b\left(h_{t}\right)\right)\nabla_{\theta}\log p_{G}\left(X\mid h_{t},\theta\right)],
\end{align*}
where $b\left(S_{t}\right)$ is a baseline, a function that can be
estimated or learned during training. The use of a baseline does not
change the gradient expected value, but in practice, reduces its variance.
In our experiments, $b(S_{t})$ is a fixed value equivalent to the
average of computed rewards over training time.

\section{Experiments\label{sec:Experiements}}

\subsection{Baselines}

We evaluate our proposed model for both discrete (in our case, text
generation) and real-valued data (air-craft trajectory generation),
summarized in Table \ref{tab:baselines}.

\begin{table}[h]
\caption{\label{tab:baselines}Comparison Baselines}

\begin{centering}
{\small{}}%
\begin{tabular}{lcc}
 & Discrete Data & Real-valued Data\tabularnewline
 & (Text) & (Trajectories)\tabularnewline
\hline 
SeqGAN\textsuperscript{{*}} & $\checked$ & --\tabularnewline
LSTM & -- & $\checked$\tabularnewline
OptiGAN-OnlyRL & $\checked$ & --\tabularnewline
OptiGAN-OnlyGAN & $\checked$ & $\checked$\tabularnewline
OptiGAN & $\checked$ & $\checked$\tabularnewline
\end{tabular}{\small\par}
\par\end{centering}
\begin{centering}
\textit{\scriptsize{}}\textsuperscript{\textit{\scriptsize{}{*}}}\textit{\scriptsize{}SeqGAN
works only with discrete data, not real-valued data}{\scriptsize\par}
\par\end{centering}
\centering{}\vspace{-2mm}
\end{table}

For text generation, we compare with three baselines:
\begin{enumerate}
\item \textbf{SeqGAN} \cite{yu2017seqgan}: is a well-known baseline for
sequential generative models that uses a discriminator as a reward
signal for training the generator in reinforcement learning framework.
\item \textbf{OptiGAN-OnlyRL: }This model is the vanilla reinforcement learning
using policy gradients. For fairness, we implement it by using our
own model with GAN component canceled, by zeroing out the GAN loss
part.
\item \textbf{OptiGAN-OnlyGAN}: The sequence GAN with LSTM and discrete
relaxation nodes, without any policy gradient component. We implement
it using our model with RL component canceled, by zeroing out the
policy gradient loss.
\end{enumerate}
The GAN network implementation of our model is based on RELGAN with
same hyperparamters and temperature scheduling, but using LSTM unit
instead of relational memory. 

For trajectory generation, we implement two different models to compare
with; \textbf{LSTM} (The LSTM component of our model without adversarial
training) and \textbf{OptiGAN-OnlyGAN} (our model without RL component),
and we conduct ablation study for the effect of the RL component.

\subsection{Text generation}

\subsubsection{Evaluation Metrics}

We use both BLEU score and negative log-likehood (NLL) mentioned below
to evaluate the quality of our model.

\paragraph*{BLEU Score}

As discussed in Section \ref{subsec:Text-generation}, BLEU score
\cite{DBLP:conf/acl/PapineniRWZ02} is well-known text quality score
in machine translation and text generation tasks.

The higher the BLEU score is, the more the number of matching n-grams
with the test set. In practice, and as discussed later, the BLEU score
can be easily cheated by repeating few matching n-grams in one sentence,
or by generating only one or few high quality sentences from the model
after training. This situation implies low output quality or diversity
from the model.

\paragraph*{Negative Log-Likelihood (NLL)}

We use the negative log-likelihood of the generator \cite{nie2018relgan}
to measure diversity, defined as:

\[
\mathrm{NLL}_{\mathrm{gen}}=-\mathbb{E}_{x_{1:T}\sim P_{d}}\log P_{G_{\theta}}\left(x_{1},\cdots,x_{T}\right)
\]

\noindent where $P_{d}$ and $P_{G_{\theta}}$ are the real data and
generated data distributions, respectively. The lower the value, the
closer the model distribution is to the empirical data distribution.

\subsubsection{Datasets}

Two text datasets were used in our experiments for text generation
are
\begin{itemize}
\item \textbf{The MS-COCO image captions dataset }\cite{chen2015microsoft}
includes 4,682 unique words with the maximum sentence length 37. Both
the training and test data contain 10,000 text sentences.
\item \textbf{The EMNLP2017 WMT News dataset} \cite{guo2018long} consists
of 5,119 unique words with the maximum sentence length 49 after using
first 10,000 sentences from \cite{nie2018relgan}. Both the training
and test data contain 10,000 sentences.
\end{itemize}

\subsubsection{Experimental settings and results}

 For MS-COCO dataset, we use policy gradient baseline value $b\left(s_{t}\right)=2.5$
and $\alpha=2.0$ for both Vanilla-RL and our model. The number of
Mone Carlo samples we use during training is $3$. For EMNLP News,
we use $b\left(s_{t}\right)=2$ and $5$ Monte Carlo samples.

In all experiments, we use gradient clipping value of $10.0$ for
the generator. In Tables \ref{tab:score_results-MS_COCO-1} and \ref{tab:score_results-EMNLP-1}
we report the means and standard deviations of test BLEU scores and
training negative likelihoods values of our model compared to other
baselines.
\begin{table*}[t]
\begin{centering}
\caption{\label{tab:score_results-MS_COCO-1}BLEU scores and NLL values on
MS-COCO dataset}
\par\end{centering}
\begin{centering}
{\scriptsize{}}%
\begin{tabular}{lc|c|c|c|c}
 & {\scriptsize{}BLEU-2 $\uparrow$} & {\scriptsize{}BLEU-3} & {\scriptsize{}BLEU-4} & {\scriptsize{}BLEU-5} & {\scriptsize{}NLL $\downarrow$}\tabularnewline
\hline 
{\scriptsize{}SeqGAN} & {\scriptsize{}$75.09\pm0.84$} & {\scriptsize{}$51.58\pm1.06$} & {\scriptsize{}$32.06\pm0.98$} & {\scriptsize{}$20.03\pm0.68$} & {\scriptsize{}$0.830\pm0.176$}\tabularnewline
\hline 
{\scriptsize{}OptiGAN-OnlyRL} & {\scriptsize{}$79.23\pm3.76$} & {\scriptsize{}$59.23\pm6.21$} & {\scriptsize{}$40.65\pm7.15$} & {\scriptsize{}$27.11\pm6.36$} & {\scriptsize{}$0.803\pm0.106$}\tabularnewline
{\scriptsize{}OptiGAN-OnlyGAN} & {\scriptsize{}$75.96\pm0.71$} & {\scriptsize{}$53.79\pm0.99$} & {\scriptsize{}$34.34\pm0.86$} & {\scriptsize{}$21.51\pm0.56$} & \textbf{\scriptsize{}$\boldsymbol{0.735\pm0.080}$}\tabularnewline
\hline 
{\scriptsize{}OptiGAN (RL+GAN)} & \textbf{\scriptsize{}$\boldsymbol{76.42\pm0.70}$} & \textbf{\scriptsize{}$\boldsymbol{54.40\pm0.99}$} & \textbf{\scriptsize{}$\boldsymbol{35.06\pm0.90}$} & \textbf{\scriptsize{}$\boldsymbol{22.25\pm0.66}$} & {\scriptsize{}$0.737\pm0.082$}\tabularnewline
\end{tabular}{\scriptsize\par}
\par\end{centering}
\centering{}\vspace{-2mm}
\end{table*}
\begin{table*}[t]
\begin{centering}
\caption{\label{tab:score_results-EMNLP-1}BLEU scores and NLL values on EMNLP
News 2017 Dataset}
\par\end{centering}
\begin{centering}
{\scriptsize{}}%
\begin{tabular}{lc|c|c|c|c}
 & {\scriptsize{}BLEU-2 $\uparrow$} & {\scriptsize{}BLEU-3} & {\scriptsize{}BLEU-4} & {\scriptsize{}BLEU-5} & {\scriptsize{}NLL $\downarrow$}\tabularnewline
\hline 
{\scriptsize{}SeqGAN} & {\scriptsize{}$\boldsymbol{76.05\pm1.67}$} & {\scriptsize{}$47.60\pm1.51$} & {\scriptsize{}$23.88\pm0.88$} & {\scriptsize{}$12.05\pm0.40$} & {\scriptsize{}$2.359\pm0.272$}\tabularnewline
\hline 
{\scriptsize{}OptiGAN-OnlyRL} & {\scriptsize{}$79.16\pm2.18$} & {\scriptsize{}$53.57\pm4.00$} & {\scriptsize{}$31.26\pm4.66$} & {\scriptsize{}$16.67\pm3.14$} & {\scriptsize{}$2.267\pm0.154$}\tabularnewline
{\scriptsize{}OptiGAN-OnlyGAN} & {\scriptsize{}$73.15\pm2.35$} & {\scriptsize{}$48.00\pm1.32$} & {\scriptsize{}$26.12\pm1.00$} & {\scriptsize{}$13.77\pm0.71$} & {\scriptsize{}$2.234\pm0.152$}\tabularnewline
\hline 
{\scriptsize{}OptiGAN (RL+GAN)} & {\scriptsize{}$74.03\pm1.69$} & {\scriptsize{}$\boldsymbol{48.73\pm1.08}$} & {\scriptsize{}$\boldsymbol{26.64\pm1.07}$} & {\scriptsize{}$\boldsymbol{14.05\pm0.79}$} & {\scriptsize{}$\boldsymbol{2.226\pm0.148}$}\tabularnewline
\end{tabular}{\scriptsize\par}
\par\end{centering}
\centering{}\vspace{-2mm}
\end{table*}

\textbf{Quality and diversity discussion}

Tables \ref{tab:score_results-MS_COCO-1} and \ref{tab:score_results-EMNLP-1}
show that, except for the OptiGAN-OnlyRL special case, our model outperforms
the baselines in BLEU scores on MS-COCO dataset and all but BLEU-2
for EMNLP News dataset. Our model also achieves a competitive NLL
value with the best model, OptiGAN-OnlyGAN. This means that our model
does not sacrifice the diversity of generated output when optimizing
for the given score. We find that SeqGAN suffers the worst NLL score,
even when compared to OptiGAN-OnlyRL. Since SeqGAN modified generator
objective does not encourage matching the model distribution to data
distribution, it can be susceptible to diversity loss. On the other
hand, GANs that use Gumbel-Softmax to keep the standard generator
objective, like ours, are more able to match the model to data distribution.

\begin{figure}
\begin{centering}
\includegraphics[width=0.82\columnwidth]{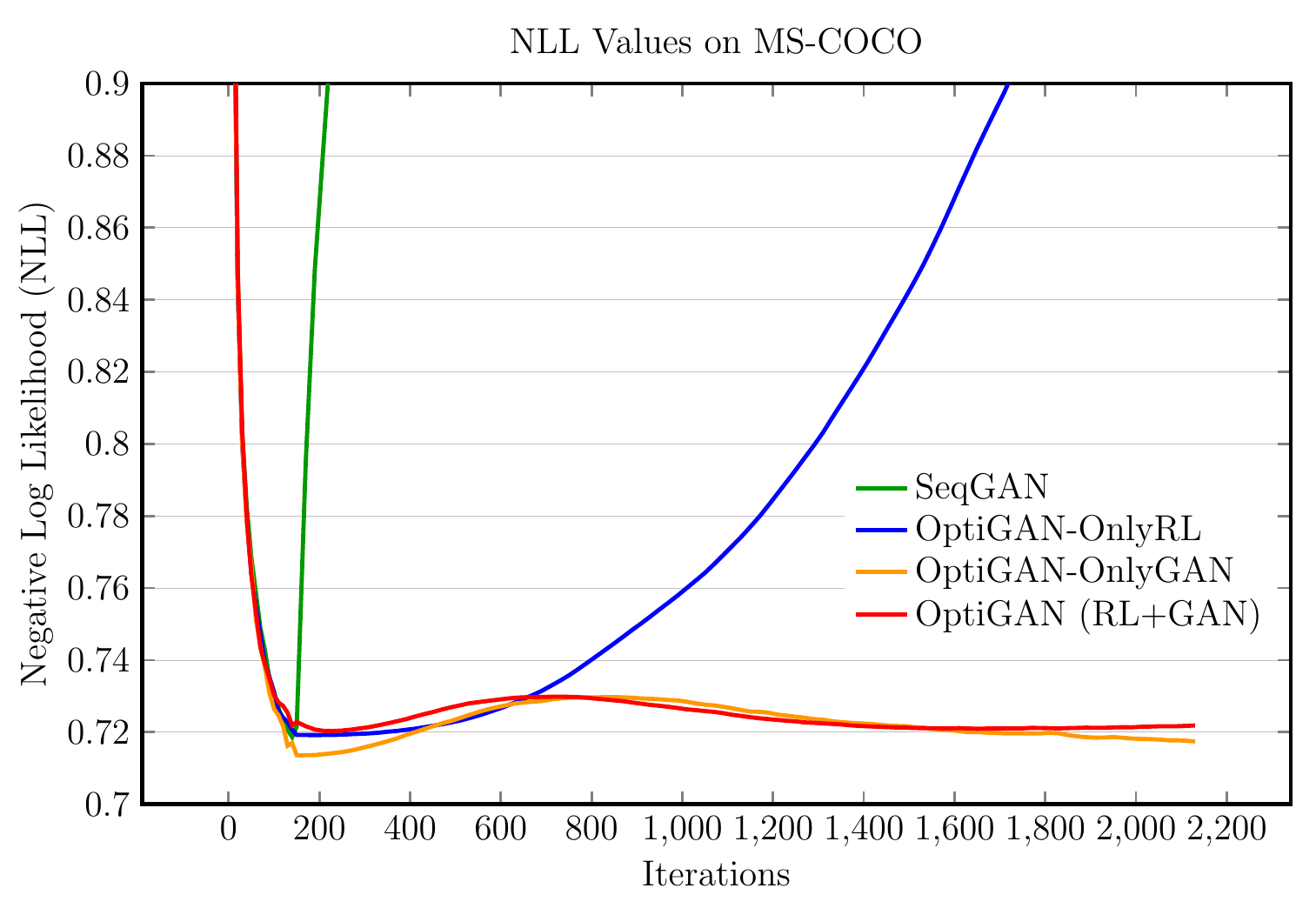}
\par\end{centering}
\centering{}\caption{\label{fig:NLL-Values} NLL values on MS-COCO Dataset. Unlike SeqGAN,
OptiGAN does not sacrifice output diversity.}
\end{figure}

In the case of OptiGAN-OnlyRL, we find that pure reinforcement learning
can achieve a higher BLEU score than other models (with very high
variance). However, it has worse NLL values, which means it has worse
diversity than our model. Fig. \ref{fig:NLL-Values} shows that OptiGAN-OnlyRL
fails to converge to low NLL, unlike our model, which has competitive
NLL values with OptiGAN-OnlyGAN. 

Moreover, although pure reinforcement learning can reach high BLEU
scores, yet the sentences mostly are not realistic. We show in Table
\ref{tab:Generated-Sample-Sentences_RL_only} sentences from OptiGAN-OnlyRL,
where we find that many of the generated sentences are unrealistic
repetitions of certain n-grams in the test set. In the case of MS-COCO
dataset, the generated sentences lengths are shorter than the average
length of the dataset. This behavior possibly means that in the absence
of the GAN objective part of the loss, pure reinforcement learning
does not have incentive to generate sentences close to the real data
distribution. In this case, the model only has to achieve high BLEU
score to reduce the optimization loss. 

\begin{table}
\caption{\label{tab:Generated-Sample-Sentences_RL_only}Sentences from OptiGAN-OnlyRL.
Pure RL looses structure with BLEU rewards, repeating certain n-grams}
\medskip{}

\centering{}{\footnotesize{}}%
\begin{tabular*}{1\columnwidth}{@{\extracolsep{\fill}}l}
\addlinespace
\textcolor{black}{\scriptsize{}Generated sentences from }\textbf{\scriptsize{}OptiGAN-OnlyRL}\tabularnewline
\midrule
\addlinespace
\textcolor{black}{\scriptsize{}the party has pledged a plate coach
and don ' t think it was good to the real head .}\tabularnewline
\midrule 
\addlinespace
\textcolor{black}{\scriptsize{}and i ' ve been - it }\textcolor{black}{\scriptsize{}\uline{'
i ' i ' i}}\textcolor{black}{\scriptsize{} have that thing a couple
to hear the end but }\textcolor{black}{\scriptsize{}\uline{i '}}\textcolor{black}{\scriptsize{}
m , ``}\tabularnewline
\addlinespace
\textcolor{black}{\scriptsize{}~~~it ' s really people were going
to do , ' it ' he work .}\tabularnewline
\midrule
\textcolor{black}{\scriptsize{}\uline{i ' d like}}\textcolor{black}{\scriptsize{}
, when}\textcolor{black}{\scriptsize{}\uline{ i ' d like}}\textcolor{black}{\scriptsize{}
to give them to that it , there , and it ' ll happen looking to}\tabularnewline
\addlinespace[1pt]
\textcolor{black}{\scriptsize{}~~~doubt that everybody challenges
...}\tabularnewline
\midrule
\addlinespace[1pt]
{\scriptsize{}if i ' ve got no evidence ,}{\scriptsize{}\uline{
it '}}{\scriptsize{} s}{\scriptsize{}\uline{ it '}}{\scriptsize{}
i expect ' there }\textcolor{black}{\scriptsize{}\uline{' he '
he}}{\scriptsize{} '}{\scriptsize{}\uline{ it ' he }}{\scriptsize{}'}{\scriptsize{}\uline{
it '}}{\scriptsize{} out , and}\tabularnewline
\addlinespace[1pt]
{\scriptsize{}~~~information it ' it ' ...}\tabularnewline
\midrule
\addlinespace[1pt]
{\scriptsize{}\uline{' i ' i }}\textcolor{black}{\scriptsize{}'
it ' we }{\scriptsize{}\uline{' i }}\textcolor{black}{\scriptsize{}think
}{\scriptsize{}\uline{i}}\textcolor{black}{\scriptsize{} ' we spent
a escape }{\scriptsize{}\uline{i ' i '}}\textcolor{black}{\scriptsize{}
ve been on the}{\footnotesize{} ...}\tabularnewline
\addlinespace[1pt]
\end{tabular*}{\footnotesize\par}
\end{table}

\textbf{Sentences quality of OptiGAN}

Table \ref{tab:Generated-Sample-Sentences_Ours} shows generated sentences
of OptiGAN. The sentences generally look meaningful, structured and
diverse, hence showing the capacity of OptiGAN in generating good
and diverse sentences.

\begin{table}
\caption{\label{tab:Generated-Sample-Sentences_Ours}Generated sample sentences
from our model}
\medskip{}

\centering{}{\footnotesize{}}%
\begin{tabular*}{1\columnwidth}{@{\extracolsep{\fill}}l}
\textbf{\scriptsize{}Samples from MS-COCO}\tabularnewline
\midrule
{\scriptsize{}a roadside vendor sells food to passersby on there
are two multitcolored towels .}\tabularnewline
\midrule
{\scriptsize{}an older man sitting at a kitchen with stainless steel
appliances .}\tabularnewline
\midrule
{\scriptsize{}a woman standing in a field with mountains in the view
of a field and a bus stop .}\tabularnewline
\midrule
{\scriptsize{}a clean bathroom with a blue toilet .}\tabularnewline
\midrule
{\scriptsize{}a group of people is watching buses next to a tall building
.}\tabularnewline
\midrule
{\scriptsize{}a woman in a white shirt and jeans walking up a air
gondola wears a}\tabularnewline
{\scriptsize{}~~~costume decorations in a red jacket hides building}\tabularnewline
\midrule
{\scriptsize{}three small dogs under a towel rack .}\tabularnewline
\midrule
{\scriptsize{}a bathroom with a toilet and a large mirror}\tabularnewline
\midrule
{\scriptsize{}a city street with cars vehicles parked on the ground
.}\tabularnewline
\midrule
{\scriptsize{}a large passenger jet flying through the air flying
a kite and an airport .}\tabularnewline
\addlinespace
\textbf{\scriptsize{}Samples from EMNLP News}\tabularnewline
\midrule
{\scriptsize{}people had gone in a few weeks ago , it ' s really
very quiet tonight to do .}\tabularnewline
\midrule
\addlinespace[1pt]
{\scriptsize{}she is me but it ' s a concept , the girls can build
high strength .}\tabularnewline
\midrule
\addlinespace[1pt]
{\scriptsize{}i ' ve got a shock for their parents law to the same
offence .}\tabularnewline
\midrule
\addlinespace[1pt]
{\scriptsize{}they ' re going to acknowledge that their football leader
will be able to get}\tabularnewline
\addlinespace[1pt]
~~~{\scriptsize{}every most republicans .}\tabularnewline
\midrule
\addlinespace[1pt]
{\scriptsize{}a tory source said : ' the 22 fall in the family in
all of the newcastle day .}\tabularnewline
\midrule
\addlinespace[1pt]
{\scriptsize{}at cbs, that is more difficult, this is that the social
stuff isn't quite any tribute for britain}\tabularnewline
\midrule
\addlinespace[1pt]
{\scriptsize{}it ' s a safe model from making a book of the first
lady who had to respond .}\tabularnewline
\addlinespace[1pt]
\end{tabular*}{\footnotesize\par}
\end{table}

\subsection{Air-Combat Trajectory Generation\label{subsec:Air-Combat-Maneuver-Trajectory-1}}

\subsubsection{Evaluation metrics}

We use the McGrew score\cite{mcgrew2010air} which measures how good
is the aircraft positioned in an attempt to get behind the other aircraft.
McGrew score is well-known by domain experts in air-combat maneuvers.

\subsubsection{Datasets}

For trajectory generation task, we used simulated data from\textit{
ACE-Zero} simulator \cite{ijcai2017-778}. We created simulated trajectory
data for the Stern Conversion maneuver \cite{austin1990game} (Fig.
\ref{fig:Stern-Conversion-Maneuvre}) with two fighters; the blue
and the red. We created 6,000 trajectories under this scenario \textcolor{black}{}\footnote{\textcolor{black}{All trajectory data is available at \url{https://bit.ly/33k1AkT}}}.Each
trajectory contains 16 featuresfor each of the two fighters. 

\subsubsection{Experimental settings and results}

\begin{figure}[h]
\begin{centering}
\includegraphics[viewport=57.624bp 48.0065bp 1272.529bp 840.113bp,clip,width=0.95\columnwidth]{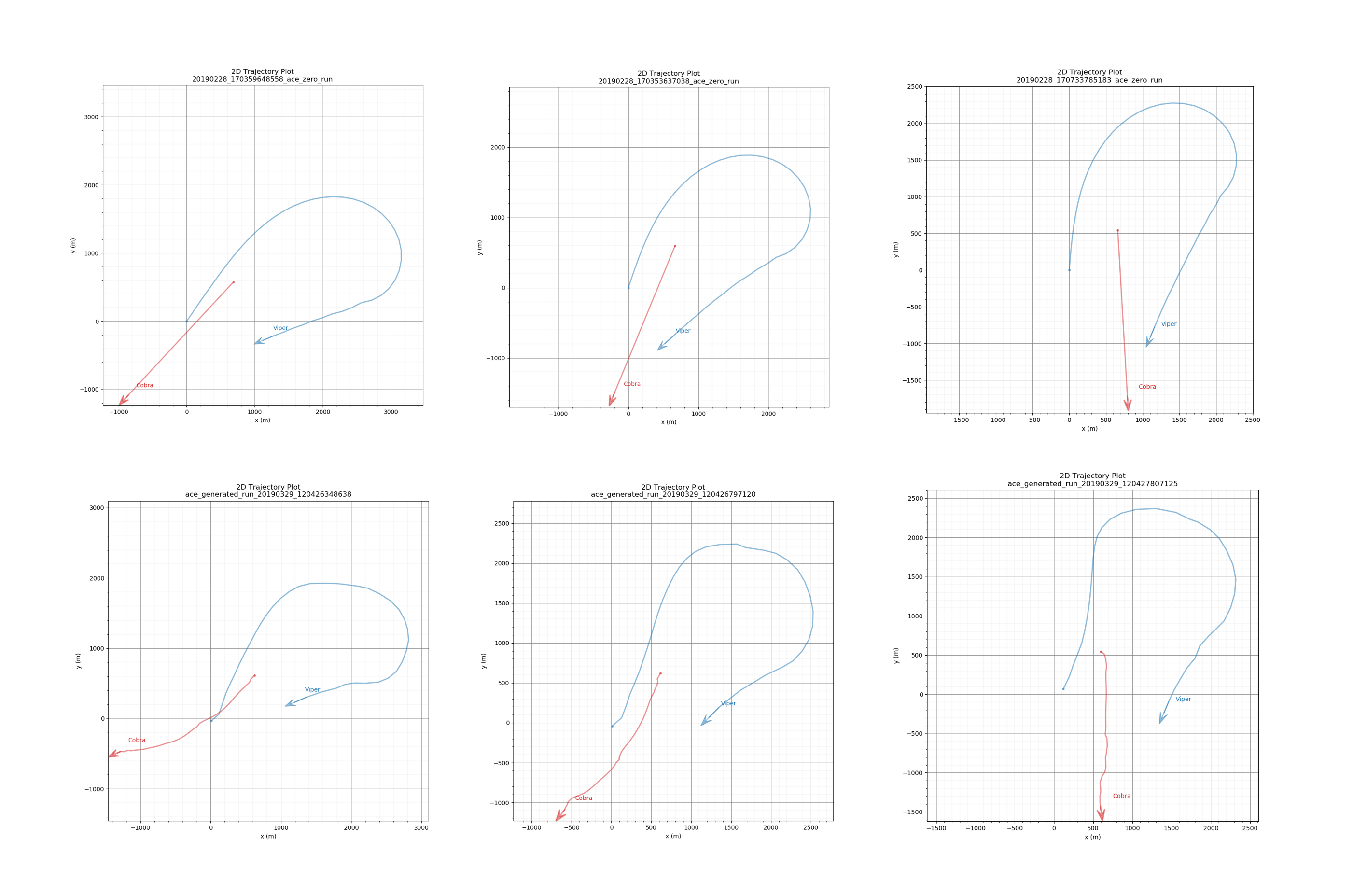}
\par\end{centering}
\caption{Samples of the training data and generated trajectories from the model.
\textit{Top row: }samples from the training trajectories in 2D position
plane. \textit{Bottom row:} generated trajectories from the trained
model (McGrew score $=6.03$ ).\label{fig:Samples_trajectories}}

\end{figure}

In all of our experiments, we use 40 simulation time steps (tokens)
for each fighter trajectory. We use 256 units hidden layer for LSTM
unit with 2 hidden layers. For the VAE part of the model, we use 12
hidden units and latent dimension of size 10. We pretrained the generator
for 80 epochs before starting the adversarial and policy gradients
training. In all experiments we set $\sigma=0$ for sampling $x_{t}$.
We show samples of the training data trajectories and generated trajectories
by our model in Fig. \ref{fig:Samples_trajectories}. We can see from
the generated trajectories is that the model is able to capture the
correct behavior, were the blue trajectory tries to get behind the
red aircraft.

\begin{table}[h]
\caption{\label{tab:score_results}Blue Fighter Engagement Scores (McGrew Score)}

\begin{centering}
{\small{}}%
\begin{tabular}{llll}
 & $\lambda$ & $\alpha$ & McGrew Score\tabularnewline
\hline 
SeqGAN\textsuperscript{{*}} &  &  & N/A\tabularnewline
\hline 
LSTM & -- & -- & 6.21\tabularnewline
OptiGAN-OnlyGAN & 1.0 & -- & 7.34\tabularnewline
OptiGAN & 0.2 & 0.75 & \textbf{8.41}\tabularnewline
\hline 
\textit{ACE0 Simulator Dataset} & \textit{--} & \textit{--} & \textit{8.53}\tabularnewline
\end{tabular}{\small\par}
\par\end{centering}
\begin{centering}
\textit{\scriptsize{}}\textsuperscript{\textit{\scriptsize{}{*}}}\textit{\scriptsize{}SeqGAN
works only with discrete data, not real-valued data}{\scriptsize\par}
\par\end{centering}
\centering{}\vspace{-2mm}
\end{table}

\begin{table}[h]
\caption{\label{tab:trajectory_ablation}Effect of hyper-paramter $\lambda$
for GAN-Only training}

\centering{}{\small{}}%
\begin{tabular}{lll}
 & $\lambda$ & McGrew Score\tabularnewline
\hline 
\multirow{2}{*}{OptiGAN-OnlyGAN} & 1.0 & 7.34\tabularnewline
\cline{2-3} \cline{3-3} 
 & 0.2 & 6.79\tabularnewline
\end{tabular}{\small\par}
\end{table}

\textbf{Score optimization}

We want the generated trajectories to be more optimized towards better
engagement position against the red fighter. The desired outcome is
a higher McGrew score, which means better engagement positions along
the generated trajectory. We evaluate the effect of using policy gradients
on the McGrew score of the blue aircraft and show the results in Table
\ref{tab:score_results}. In all experiments, we use $\gamma=0.9$.

We compare with three baselines, Our model for real-valued data without
adversarial training or policy gradients (\textbf{LSTM}), GAN without
policy gradients (\textbf{OptiGAN-OnlyGAN}), and the average McGrew
score of the training data (from simulator).

In all baselines, we generate 6,000 trajectories. . We can see that
full OptiGAN model with the policy gradients achieve higher McGrew
scores than other baselines, and closest to the real physics simulator.
Although the GAN without policy gradients was able to achieve a slightly
less score, the policy gradient model was run with the small value
$\lambda=0.2$. This means that the adversarial training did not contribute
to the high score achieved by policy gradients model, rather, it was
mainly the effect of policy gradients. As shown in Table \ref{tab:trajectory_ablation},
GAN with no PG model with $\lambda=0.2$ did not achieve the same
score as the the one with $\lambda=1.0$.

\textbf{Trajectories quality of OptiGAN}

Fig. \ref{fig:Samples_trajectories} shows the trajectories generated
by OptiGAN compared to real trajectories. It can be observed that
OptiGAN can generate high-quality trajectories resembling real data.

\section{Conclusion\label{sec:Conclusion}}

In this paper we presented a sequential deep generative model, OptiGAN,
that integrates both generative adversarial networks and reinforcement
learning for goal optimized generation. In many applications, goal
optimization is a useful mechanism to give desired properties to generated
outputs. We applied our model to text and air-combat trajectory generation
tasks, and showed that the model generated high quality sentences
with higher desired scores. In addition, OptiGAN preserves the diversity
of outputs close to the real data. Our model serves as a general framework,
that can be used for any GAN model to enable it to directly optimize
a desired goal according to the given task.

In future work, we plan to improve the quality of real-valued outputs
(e.g. trajectories) to make it more realistic to the data and physically
constrained. We further look to incorporate latent space in the discrete
case, which can be leveraged to guide the generation process along
learned disentangled features of the data.

\section*{Acknowledgments}

We gratefully acknowledge the partial support from the Defence Science
and Technology (DST) Group for this work.

\appendix

\section{Theoretical Proofs }

\subsection{{\footnotesize{}Proof of the final objective function \label{subsec:Proof-of-final-1}}}

{\footnotesize{}Consider this optimization problem:
\begin{align}
\max_{G}\min_{D}\, & \biggl[\mathbb{E}_{X\sim p_{d}}\left[\log\,p_{G}\left(X\mid\theta\right)\right]-\label{eq:op_ARN-1-1}\\
 & \text{\ensuremath{\mathbb{E}}}_{X\sim p_{d}}\left[\log\,D\left(X\right)\right]-\mathbb{E}_{z\sim p_{z}}\left[\log\left[1-D\left(G\left(z\right)\right)\right]\right]\biggr].\nonumber 
\end{align}
Given a generator $G$, the optimal $D^{*}\left(G\right)$ is determined
as:
\[
D_{G}^{*}\left(X\right)=\frac{p_{d}\left(X\right)}{p_{G}\left(X\right)+p_{d}\left(X\right)},
\]
where $p_{G}\left(X\right)$ is the distribution induced from $G\left(X\right)$
where $X\sim p_{d}\left(X\right)$.}{\footnotesize\par}

{\footnotesize{}Substituting $D_{G}^{*}$ back to Eq. (\ref{eq:op_ARN-1-1}),
we obtain the following optimization problem regarding $G$:
\begin{equation}
\max_{G}\left(\mathbb{E}_{p_{d}}\left[\log\,p_{G}\left(X\right)\right]-I_{JS}\left(P_{d}\Vert P_{G}\right)\right).\label{eq:ARN_G-1}
\end{equation}
}{\footnotesize\par}

{\footnotesize{}The objective function in Eq. (\ref{eq:ARN_G-1})
can be written as
\begin{gather*}
\mathbb{E}_{p_{d}}\left[\log\,p_{G}\left(X\right)\right]-I_{JS}\left(P_{d}\Vert P_{G}\right)\\
\,\,\,\,\,\,\,=-I_{JS}\left(P_{d}\Vert P_{G}\right)-I_{KL}\left(P_{d}\Vert P_{G}\right)-\mathbb{E}_{p_{d}}\left[\log\,p_{d}\left(X\right)\right]\\
=-I_{JS}\left(P_{d}\Vert P_{G}\right)-I_{KL}\left(P_{d}\Vert P_{G}\right)+\text{const}.
\end{gather*}
}{\footnotesize\par}

{\footnotesize{}Therefore, the optimization problem in Eq. (\ref{eq:ARN_G-1})
is equivalent to:
\[
\min_{G}\left(I_{JS}\left(P_{d}\Vert P_{G}\right)+I_{KL}\left(P_{d}\Vert P_{G}\right)\right).
\]
}{\footnotesize\par}

{\small{}At the Nash equilibrium point of this game, we hence obtain:
$p_{G}\left(X\right)=p_{d}\left(X\right)$.}{\small\par}

\bibliographystyle{IEEEtran}
\bibliography{ref_duplicate}

\end{document}